\documentclass[conference]{IEEEtran}
\ifCLASSINFOpdf
\else
\fi

\usepackage{amssymb}
\usepackage{graphicx}
\usepackage{caption}
\usepackage{subcaption}

\usepackage{algorithm}
\usepackage{algpseudocode}

\newcommand{\argmax}{\mathop{\rm arg~max}\limits}

\begin{document}
%
\title{Asymmetric Move Selection Strategies in Monte-Carlo Tree Search: Minimizing the Simple Regret at Max Nodes}


\author{\IEEEauthorblockN{Yun-Ching Liu}
\IEEEauthorblockA{ 
Graduate School of Engineering\\
University of Tokyo\\
cipherman@logos.t.u-tokyo.ac.jp}
\and
\IEEEauthorblockN{Yoshimasa Tsuruoka}
\IEEEauthorblockA{Graduate School of Engineering\\
University of Tokyo\\
tsuruoka@logos.t.u-tokyo.ac.jp}
}


%


\maketitle

\begin{abstract}
The combination of multi-armed bandit (MAB) algorithms with Monte-Carlo tree search (MCTS) has made a significant impact in various research fields. The UCT algorithm, which combines the UCB bandit algorithm with MCTS, is a good example of the success of this combination. The recent breakthrough made by AlphaGo, which incorporates convolutional neural networks with bandit algorithms in MCTS, also highlights the necessity of bandit algorithms in MCTS. However, despite the various investigations carried out on MCTS, nearly all of them still follow the paradigm of treating every node as an independent instance of the MAB problem, and applying the same bandit algorithm and heuristics on every node. As a result, this paradigm may leave some properties of the game tree unexploited. In this work, we propose that max nodes and min nodes have different concerns regarding their value estimation, and different bandit algorithms should be applied accordingly. We develop the Asymmetric-MCTS algorithm, which is an MCTS variant that applies a simple regret algorithm on max nodes, and the UCB algorithm on min nodes. We will demonstrate the performance of the Asymmetric-MCTS algorithm on the game of $9\times 9$ Go, $9\times 9$ NoGo, and Othello.         

\end{abstract}


%
\IEEEpeerreviewmaketitle

\section{Introduction}

Monte-Carlo Tree Search (MCTS) has made a significant impact on various fields in AI, especially on the field of computer Go \cite{survey}. The key factor to the success of MCTS lies in its combination with bandit algorithms, which solves the multi-armed bandit problem (MAB) \cite{uct}. The MAB problem is a problem where the agent needs to decide whether it should act optimally based on current available information ({\it exploitation}), or gather more information at the risk of suffering losses incurred by performing suboptimal actions ({\it exploration}) \cite{optimal}.  One of the most widely used MCTS variants is the UCT algorithm, which simply applies the UCB algorithm to every node in the tree \cite{ucb}. The development of MCTS in recent years can be broadly classified into two main directions: one is the integration of knowledge learnt offline, and the other is increasing the effectiveness of the knowledge accumulated online.

The integration of offline knowledge was mainly focused on using logistic models to improve the quality of the simulations \cite{sb}\cite{mm}.  Recently, a lot of effort has been put into the training of convolutional neural networks and combining them with MCTS in computer Go \cite{cnn3}\cite{alphago}. A breakthrough was made by the program {\it AlphaGo}, which essentially combines convolutional neural networks with the PUCB bandit algorithm \cite{pucb}, and has beaten a top human professional player Lee Sedol in a five-game challenge match \cite{alphago}.    

On the other hand, various investigations in increasing the effectiveness of online knowledge have also been carried out. One of them is using various bandit algorithms with MCTS, especially the bandit algorithms that solve the {\it pure exploration} MAB problem \cite{pure}. The pure exploration MAB problem is a variant of the MAB problem. Unlike the standard MAB problem, its objective is to identify the optimal action after a number of trials, rather than accumulate as much reward as possible during those trials. The goal of the pure exploration MAB problem can be equivalently formulated as the minimization of {\it simple regret}, which is defined as the difference of the expected reward between the true optimal action and the action that has been identified as the optimal action. It has been argued that {\it simple regret bandit algorithms} might be better suited to the task of game tree search, since the ultimate goal of game tree search is to find the best possible action \cite{simple}. Therefore, various MCTS variants have been proposed based on simple regret bandit algorithms. The SR+CR scheme \cite{simple} is an MCTS algorithm that applies a simple regret bandit algorithm on the root node, and the UCB algorithm on all other nodes. The sequential halving on trees (SHOT) algorithm combines the sequential halving algorithm \cite{sh}, which is a near optimal simple regret bandit algorithm, with MCTS. The Hybrid MCTS (H-MCTS) algorithm \cite{huct} first applies the UCB algorithm on each node, and then switches to the sequential halving algorithm if the number of times a node has been visited has exceeded a predetermined threshold. The CCB-MCTS algorithm \cite{ccb} uses the improved UCB algorithm to regulate the amount of exploration performed by simple regret bandit algorithms. 


However, the paradigm of applying bandit algorithms in all MCTS variants is still essentially the same: viewing every node in the game tree as an independent instance of the MAB problem, and applying the same bandit algorithm and heuristics on every node. Although this approach allows MCTS to be applied in general domains other than game-play, it leaves certain properties of the game tree unexploited. 

The adversarial game tree consists of two types of nodes: min nodes and max nodes. Max nodes and min nodes generally represent the decision of different players in the game tree, and it is conventional knowledge in various games that the decision of which strategy to adopt, should be based on which player he or she is. For example,  in the game of Go, a komi of 6.5 is given to the Black player, that is the Black player needs to obtain at least 6.5 points more than the White player to win the game. Therefore, the Black player needs to adopt a more aggressive strategy, while the White player can play more conservatively or defensively. The same can also be observed in the game of Chess, where White is generally considered to have the initiative from the start, and hence needs to play more actively, while Black needs to solve its passivity first. Therefore, max nodes and min nodes are intrinsically different from this high-level point of view, and it would be natural to treat them differently, rather than symmetrically.

Some methods have been proposed to reflect the min-max property of game trees in MCTS, but still essentially treat max nodes and min nodes symmetrically, and apply the same heuristic on every node \cite{minmax}. The SR+CR scheme differs only the root node from other nodes, rather than between max nodes and min nodes \cite{simple}.

In this paper, we propose that max nodes and min nodes should be treated differently, and one should apply different bandit algorithms for each node type in MCTS. We will develop the {\it Asymmetric-MCTS} algorithm, which applies the UCB$_{\sqrt{\cdot}}$ algorithm on max nodes and the UCB algorithm on min nodes. We will demonstrate its performance on the game of $9\times 9$ Go, $9\times 9$ NoGo, and Othello.

\section{Preliminaries}

A key ingredient in the success of MCTS is the application of bandit algorithms. {\it Bandit algorithms} are algorithms that solve the MAB problem \cite{optimal}.

In the MAB problem, an agent faces $K$ slot machines, or ``one-armed bandits'', and the agent can choose to pull one of the slot machines at each play. The chosen slot machine will then produce a reward $r\in[0,1]$. The distribution of the reward of each slot machine is unknown to the agent.

There are two possible objectives in the MAB problem, and different types of bandit algorithms are required for achieving each objective.

\subsection{Cumulative Regret Minimization}

The goal of the conventional MAB problem is to accumulate as much reward as possible over a total of $T$ plays. The objective can be equivalently formulated as the minimization of the {\it cumulative regret}, which is defined as 
\begin{center}
	$CR_T = \sum_{t=1}^T(r^* - r_{I_t})$,
\end{center}
where $r^*$ is the expected reward of the optimal arm, and $r_{I_t}$ is the reward that the agent received by pulling arm $I_t$ on play $t$. A bandit algorithm is considered optimal if it can restrict the increase of cumulative regret to $O(\log T)$ \cite{optimal}.

\begin{algorithm}
\caption{The $UCB$ algorithm \cite{ucb}}
\label{UCB}
\begin{algorithmic}
\State {\bf Initialization}: Play each machine once.
\For{$t = 1, 2, 3, \cdots$}
	\State play arm $a_i = \argmax_{i\in K} w_i + c\sqrt{\frac{\log{t}}{t_i}}$, \\where $w_i$ is the current average reward, $t_i$ is the number of times arm $a_i$ has been sampled. 
\EndFor  
\end{algorithmic}
\end{algorithm}

The UCB algorithm \cite{ucb}, which is applied in the UCT algorithm \cite{uct}, is an optimal bandit algorithm which restricts the growth of cumulative regret to $O(\frac{K\log T}{\Delta})$, where $\Delta$ is the difference of expected reward between a suboptimal arm and the optimal arm.  

The UCB algorithm, shown in Algorithm \ref{UCB}, maintains an estimated confidence bound of the expected reward of each arm, and the algorithm simply chooses the arm that has the highest upper bound to pull at each play. The UCB algorithm estimates the confidence bound of arm $a_i$ at play $t$ as
\begin{center}
	$UCB_i = w_i \pm c_r\sqrt{\frac{\log{t}}{t_i}}$,
\end{center}  
where $w_i$ is the average reward received from $a_i$ so far, and $t_i$ is the number of times $a_i$ has been played up to play $t$, and $c_r$ is a constant. It can be observed that the confidence bound consists of the {\it exploitation term} $w_i$, and {\it the exploration term} $c\sqrt{\frac{\log t}{t_i}}$. The width of the confidence bound is determined by the exploration term, and it gradually decreases as the number of times arm $a_i$ is played increases, e.g. as $t_i$ increases, the bound becomes tighter.

\subsection{Simple Regret Minimization} 

The objective of the {\it pure exploration} MAB problem is to identify the arm that has the highest expected reward after a given total amount of $T$ plays \cite{pure}. This task can be formally stated as minimizing the {\it simple regret}, which is defined as
\begin{center}
	$SR_T = r^* - r_T$,
\end{center}
where $r^*$ is the expected reward of the optimal arm, and $r_T$ is the mean reward of the arm that is identified by the agent to be optimal after the $T$ plays.  

Since the goal is to identify which arm is the optimal arm, it is more critical to gather as much information as possible about each arm, and therefore the amount of accumulated reward during these $T$ plays is irrelevant. It has been shown that the minimization of cumulative regret $CR_T$ and the minimization of simple regret $SR_T$ are two contradicting objectives, i.e., as $CR_T$ decreases, $SR_T$ will increase at the same time, and vice versa \cite{pure}. Therefore, in order to solve the pure exploration MAB problem, a different type of bandit algorithms is needed. 

\begin{algorithm}
\caption{The $UCB_{\sqrt{\cdot}}$ algorithm \cite{simple}}
\label{UCBsqrt}
\begin{algorithmic}
\State {\bf Initialization}: Play each machine once.
\For{$t = 1, 2, 3, \cdots$}
	\State play arm $a_i = \argmax_{i\in K} w_i + c\sqrt{\frac{\sqrt{t}}{t_i}}$, \\where $w_i$ is the current average reward, $t_i$ is the number of times arm $a_i$ has been sampled. 
\EndFor  
\end{algorithmic}
\end{algorithm}

The UCB$_{\sqrt{\cdot}}$ algorithm, shown in Algorithm \ref{UCBsqrt}, is a bandit algorithm that restricts the growth of simple regret to $O((\Delta \exp(-\sqrt{T}))^K)$ \cite{simple}.

The algorithmic aspect of the UCB$_{\sqrt{\cdot}}$ algorithm is basically the same as the UCB algorithm, as it also maintains an estimated confidence bound for each arm, and chooses the arm with the highest upper bound at each play.  The UCB$_{\sqrt{\cdot}}$ algorithm defines the confidence bound for arm $a_i$ as
\begin{center}
	$UCB_{i} = w_i \pm c_s\sqrt{\frac{\sqrt{t}}{t_i}}$,
\end{center} 
where $w_i$ is the average reward received so far from arm $a_i$, $t_i$ is the number of times that $a_i$ has been played up to play $t$, and $c_s$ is a constant. As with the UCB algorithm, the confidence bound of the UCB$_{\sqrt{\cdot}}$ algorithm also consists of the {\it exploitation term} $w_i$ and the {\it exploration term} $c\sqrt{\frac{\sqrt{t}}{t_i}}$. 

The difference between the UCB and the UCB$_{\sqrt{\cdot}}$ algorithm lies in the definition of the exploration term. The exploration term for the UCB$_{\sqrt{\cdot}}$ algorithm decreases more slowly than that of the UCB algorithm, and hence tends to sample more arms over time than focusing on the arm that currently seems to be optimal.

\section{Asymmetric Monte-Carlo Tree Search}

MCTS consists of four major steps: {\it selection}, {\it expansion}, {\it simulation}, and {\it backpropagation}. Bandit algorithms are mainly applied in the {\it selection} phase by viewing each node as an independent instance of the MAB problem, where each child node is a single candidate arm. Currently, the most popular variant of MCTS is the UCT algorithm, in which the UCB algorithm is the applied bandit algorithm.

Although this general MCTS paradigm allows it to be applied in a wide range of domains, it leaves a number of properties of the game tree unexploited. 

\begin{figure}
	\centering
	\caption{An example tree for which the UCT algorithm has very poor performance \cite{lower}.}
	\label{fig:tree}
	\includegraphics[scale=0.3]{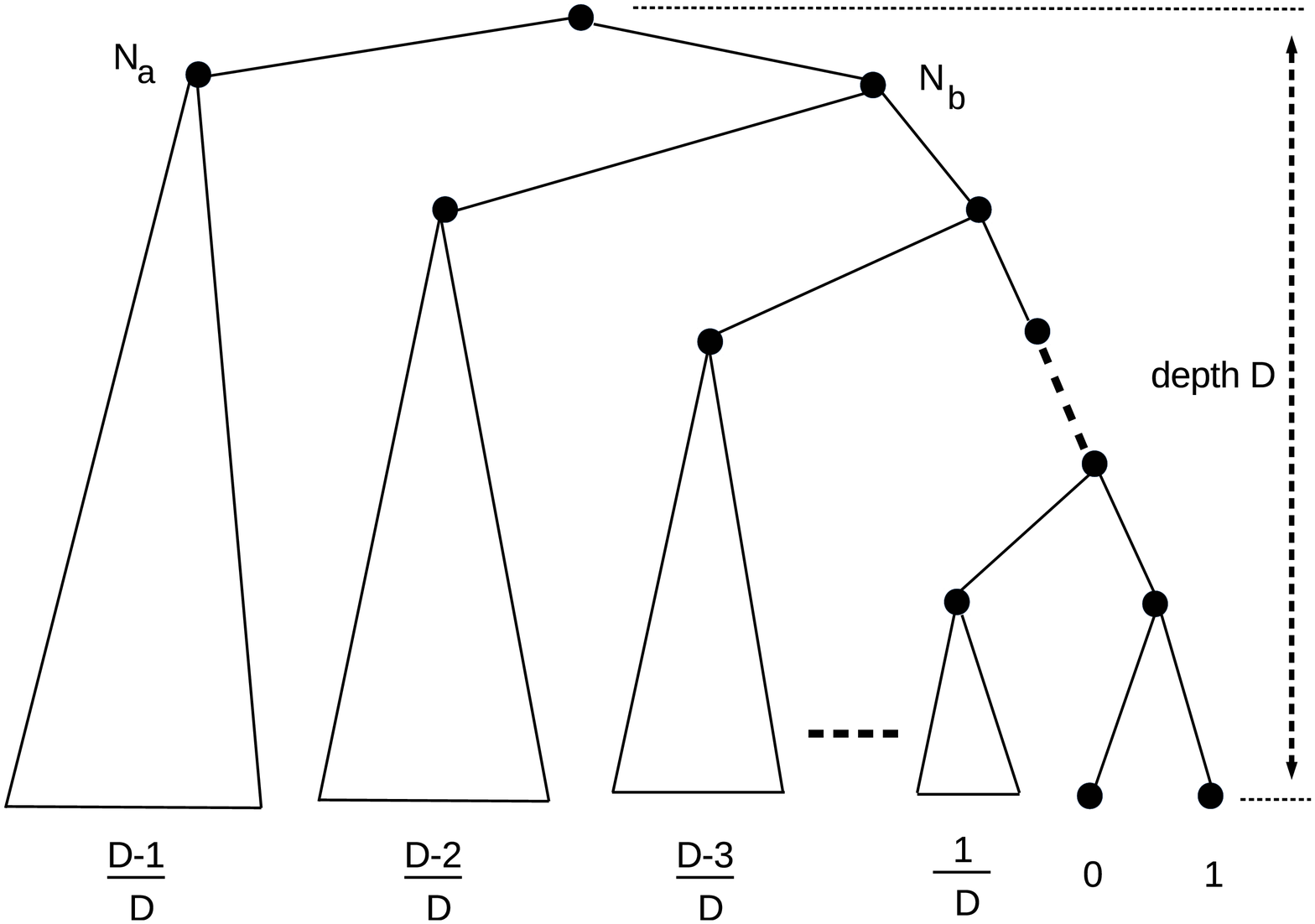}
\end{figure}

\begin{figure}
	\centering
	\caption{Asymmetric-MCTS algorithm. The gray nodes are max nodes, which the UCB$_{\sqrt{\cdot}}$ bandit algorithm are applied, and the white nodes are min nodes, which the UCB bandit algorithm applied.}
	\label{fig:nodewise}
	\includegraphics[scale=0.45]{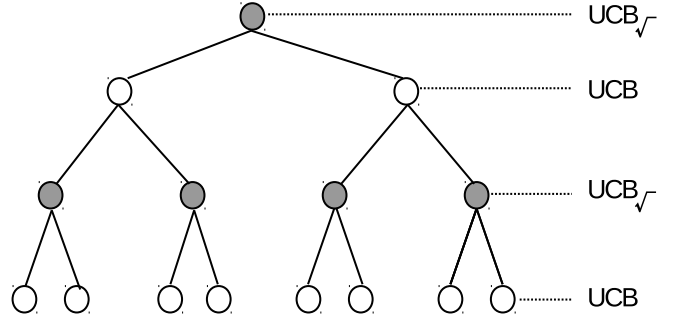}
\end{figure}

\subsection{Concerns on Value Estimation of Different Node Types}

The role of the bandit algorithm on every node of MCTS is to estimate the value of the node and perform selection according to the estimated value. As the search progresses, the estimation value of the nodes also converges. Although the general goal is to obtain a good estimation as fast as possible, it can be observed that different node types in the game tree have different requirements to their estimated values:

\begin{itemize}
	\item {\bf Max node}: since the max nodes represent the view point of the current decision maker, we need to be more certain about the estimated value of each possible decision. Estimations should also be more cautious, and not overly optimistic. 
	
	\item {\bf Min node}: since the min nodes represent the reaction of the opponent, it is not necessary to determine the best possible reaction of the opponent. Just a {\it good enough} reaction that is sufficient to refute a decision made by the decision maker will do. 
\end{itemize}

\begin{algorithm}
\caption{Asymmetric-MCTS Algorithm}
\label{NodeMCTS}
\begin{algorithmic}
\Function {Asymmetric-MCTS}{Node $N$}
\State $best_{ucb} \gets -\infty$
\For{all child nodes $n_i$ of $N$}
	\If{$n_i.t = 0$}
		\State $n_i.ucb \gets \infty$
	\Else
		\If{$N.type$ is $MAX$}
			\State $n_i.ucb \gets n.w + c_s\cdot \sqrt{\frac{\sqrt{N.t}}{n_i.t}}$
		\Else
			\State $n_i.ucb \gets n.w + c_r\cdot \sqrt{\frac{\log N.t}{n_i.t}}$
		\EndIf
	\EndIf
	\If{$best_{ucb} \leq n_i.ucb$}
		\State $best_{ucb}\gets n_i.ucb$
		\State $n_{best}\gets n_i$
	\EndIf
\EndFor
\State
\If{$n_{best}.t = 0$}
	\State $result \gets $\Call{RandomSimulation}{$n_{best}$}
\Else
	\State {\bf if} $n_{best}$ is not expanded {\bf then} \Call{Expand}{$n_{best}$}  
	\State $result \gets $ \Call{Asymmetric-MCTS}{$n_{best}$}
\EndIf
\State 
\State $n_{best}.w \gets (n_{best}.w \times n_{best}.t + result) / (n_{best}.t + 1)$
\State $n_{best}.t \gets n_{best}.t + 1$
\State $N.t \gets N.t + 1$
\State
\Return $result$
\EndFunction 
 
\end{algorithmic}
\end{algorithm}

Due to the selection and expansion performed in MCTS, the reward of the MAB problem at each node is non-stationary \cite{lower}. For example, consider the binary tree used for constructing a lower bound for the UCT algorithm \cite{lower}, shown in Fig. \ref{fig:tree}. The binary tree has the depth of $D$, and the rightmost path, which is from the root node to the rightmost leaf node, is the optimal path. For a node $N$ at depth $d<D$ on the optimal path, if the left action is chosen, then a reward of $\frac{D-d}{D}$ is received. In other words, all the leaf nodes of the subtree rooted at $N$ have the value of $\frac{D-d}{D}$. If the right action is chosen, the agent can proceed to expand further down the optimal path. At depth $D-1$ of the optimal path, the left action will give the reward $0$, and the right action will give the reward $1$. Therefore, MCTS will most likely spend the majority of its time expanding the subtrees of the left action along the optimal path, as it seems to be better. Consider the MAB problem at the root node, which has two arms node $N_a$ and node $N_b$. Since the leaf nodes of the subtree rooted at $N_a$ all have the value of $\frac{D-1}{D}$, the reward produced by $N_a$ will most likely be fixed around $\frac{D-1}{D}$. However, as the search gradually expand down the optimal path, the reward produced by $N_b$ will most likely be along the sequence
\begin{center}
	$\{ \frac{D-2}{D},\cdots, \frac{D-2}{D}, \frac{D-3}{D},\cdots, \frac{D-3}{D}, \cdots, \frac{1}{D},\cdots,\frac{1}{D}, 1\}$,
\end{center}
instead of being more evened out. Therefore, although the distribution of the reward of the MAB problem on each node is fixed and determined by the values of the leaf node, due to the selection and expansion performed in MCTS, the reward of the MAB problem on each node is biased, and hence affect the estimation made by the bandit algorithms.  

Consider the case where a sequence of rewards $(r_1, r_2, r_3, \cdots, r_n)$ are drawn from distribution $\mathcal{N}$, but due to some sampling bias, the sequence is ordered in a non-decreasing order, that is $r_i \leq r_j$ if $i < j$. Therefore, the estimated mean reward will be higher than the true mean reward in the early period of the sequence, and hence causing the agent to be too optimistic. Similarly, if the sequence is in a non-increasing order, that is $r_i \geq r_j$ if $i > j$, then the agent tends to be underestimate the mean in the early stages. 

Therefore, one should choose a bandit algorithm that is most likely to resist over optimistic estimations caused by biased reward to deploy on max nodes, and a bandit algorithm that can adapt itself rapidly to provide a ``good enough'' estimation on min nodes. 

 \begin{figure}
	\centering
	\caption{Optimal percentage of biased reward MAB problem}
    \begin{subfigure}[b]{0.5\textwidth}
        \includegraphics[width=\textwidth]{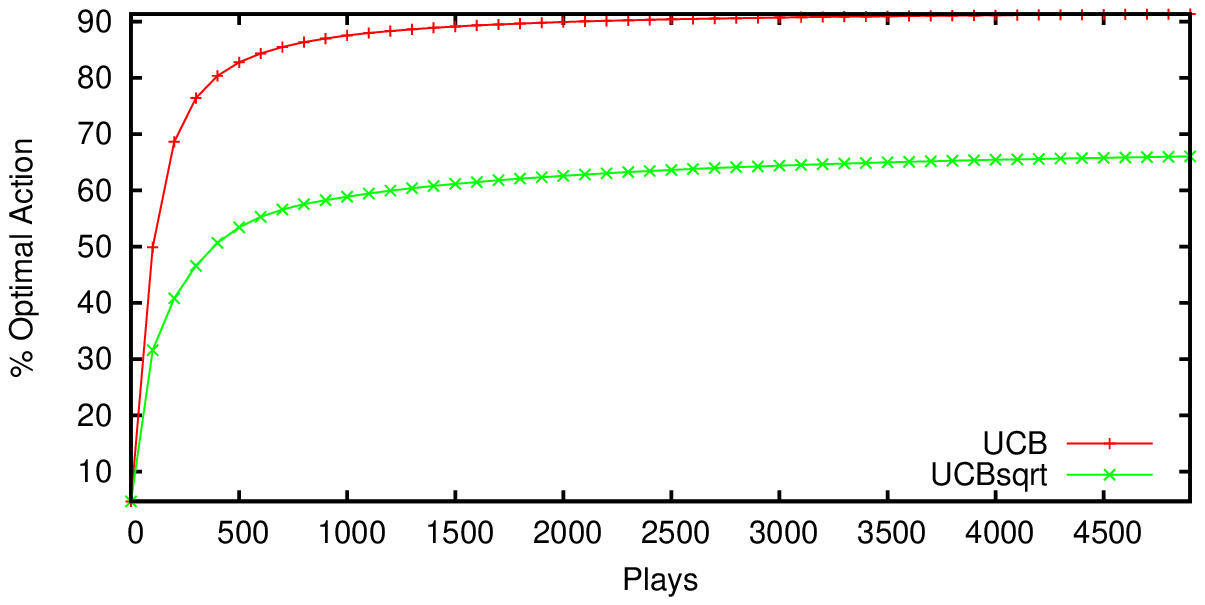}
        \caption{ascending reward}
        \label{fig:aop}
    \end{subfigure}
    ~ 
    \begin{subfigure}[b]{0.5\textwidth}
        \includegraphics[width=\textwidth]{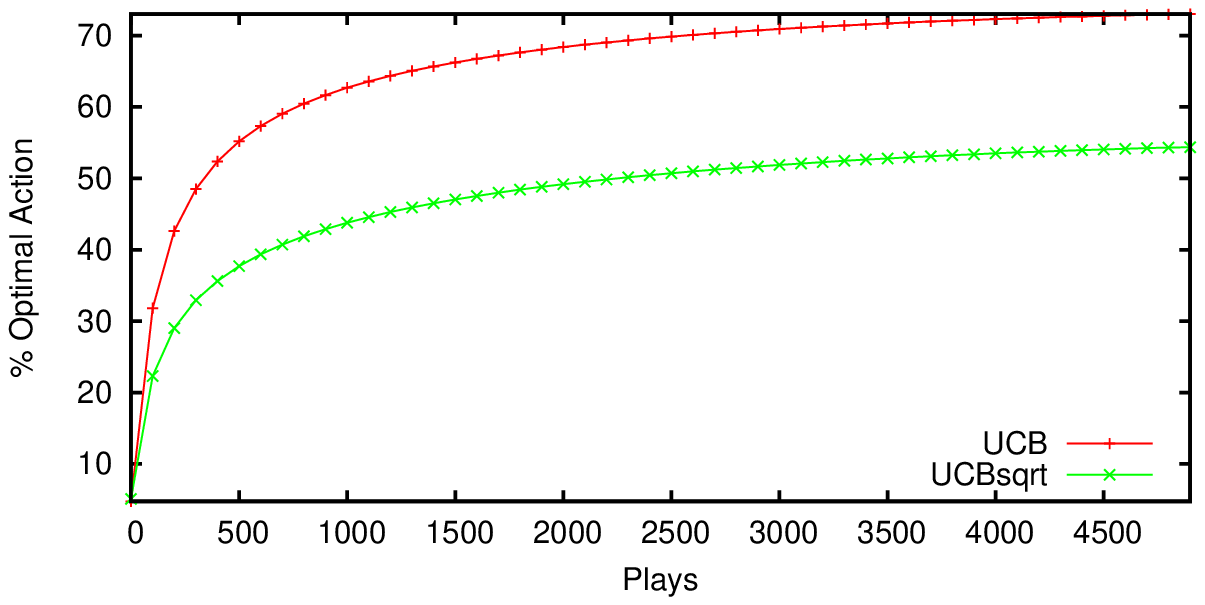}
        \caption{descending reward}
        \label{fig:dop}
    \end{subfigure}
\end{figure}

 \begin{figure}
	\centering
	\caption{Cumulative regret of biased reward MAB problem.}
    \begin{subfigure}[b]{0.5\textwidth}
        \includegraphics[width=\textwidth]{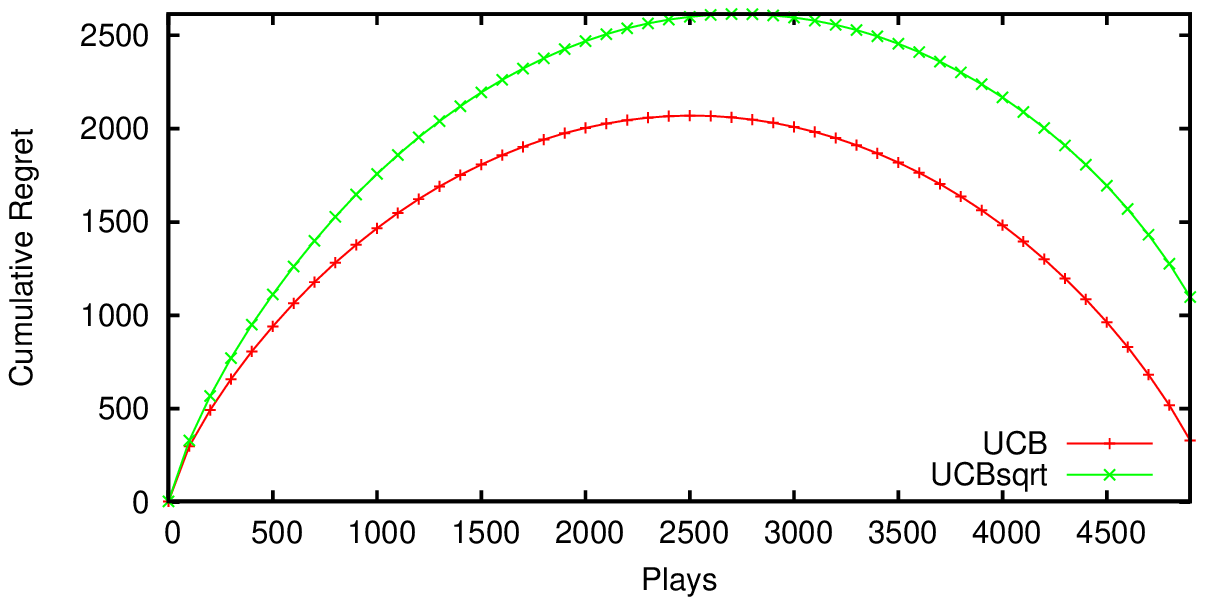}
        \caption{ascending reward}
        \label{fig:acr}
    \end{subfigure}
    ~ 
    \begin{subfigure}[b]{0.5\textwidth}
        \includegraphics[width=\textwidth]{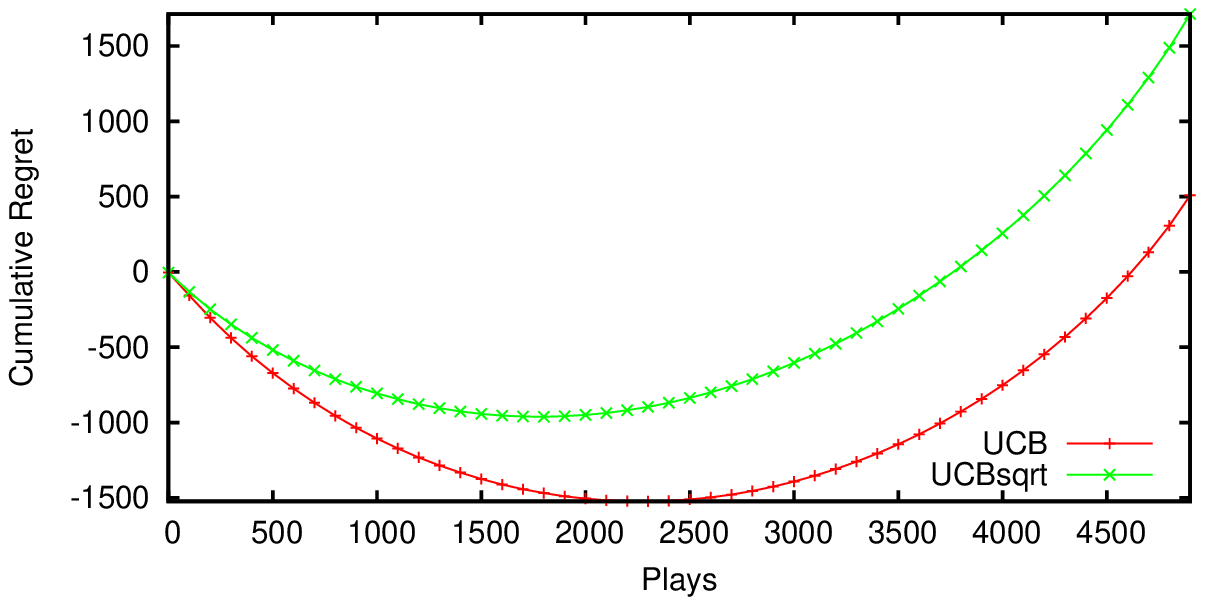}
        \caption{descending reward}
        \label{fig:dcr}
    \end{subfigure}
\end{figure} 
     
\subsection{Different Bandit Algorithms for Different Node Types}  

As simple regret and cumulative regret bandit algorithms have different properties, they can be deployed to different node types accordingly to fulfill the requirements on the estimation value of each node type:  

\begin{itemize}
	\item {\bf Max node}: {\it simple regret} bandit algorithms, which determine the optimal arm, have a higher level of confidence in its estimation value of each arm. Moreover, in order to provide a better estimation of the value of each arm, simple regret bandit algorithms tend to perform more exploration, and spread its sampling more evenly across the candidates, which effectively make it less likely to be too optimistic. 
	
	\item {\bf Min node}: {\it cumulative regret} bandit algorithms, which try to accumulate as much reward as possible, tend to focus on the current optimal arm, and adapt rapidly if the current optimal arm changes. Therefore, cumulative regret bandit algorithms seem to fit the requirement of finding a {\it good enough} reaction to refute a candidate decision. 
\end{itemize}

 \begin{figure}
	\centering
	\caption{Simple regret of biased reward MAB problem.}
    \begin{subfigure}[b]{0.5\textwidth}
        \includegraphics[width=\textwidth]{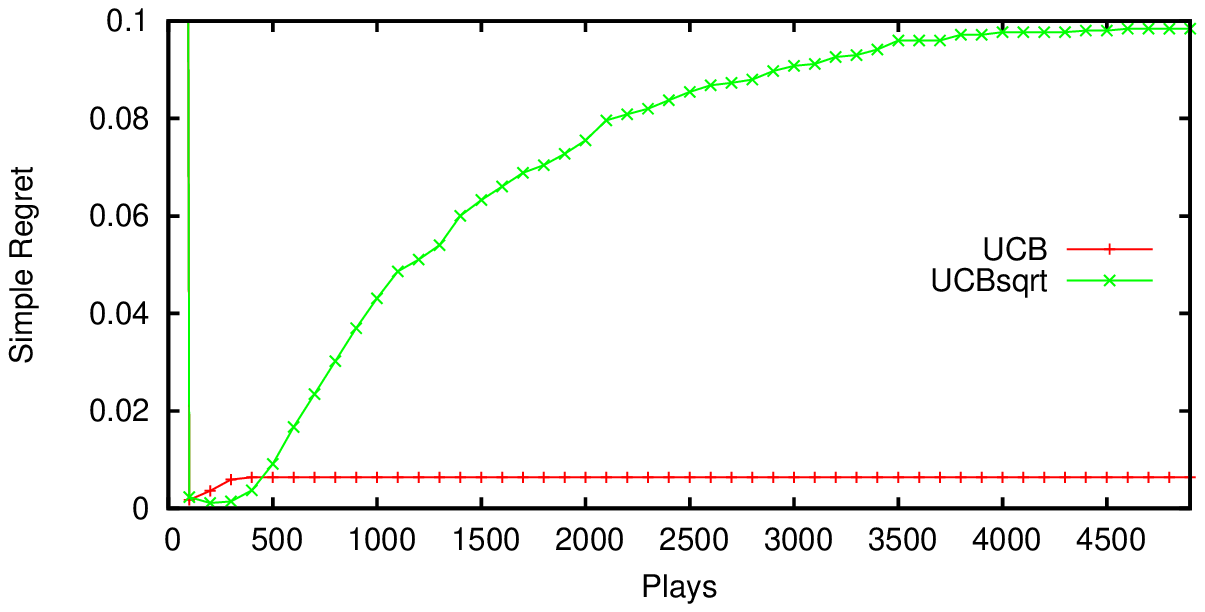}
        \caption{ascending reward}
        \label{fig:asr}
    \end{subfigure}
    ~ 
    \begin{subfigure}[b]{0.5\textwidth}
        \includegraphics[width=\textwidth]{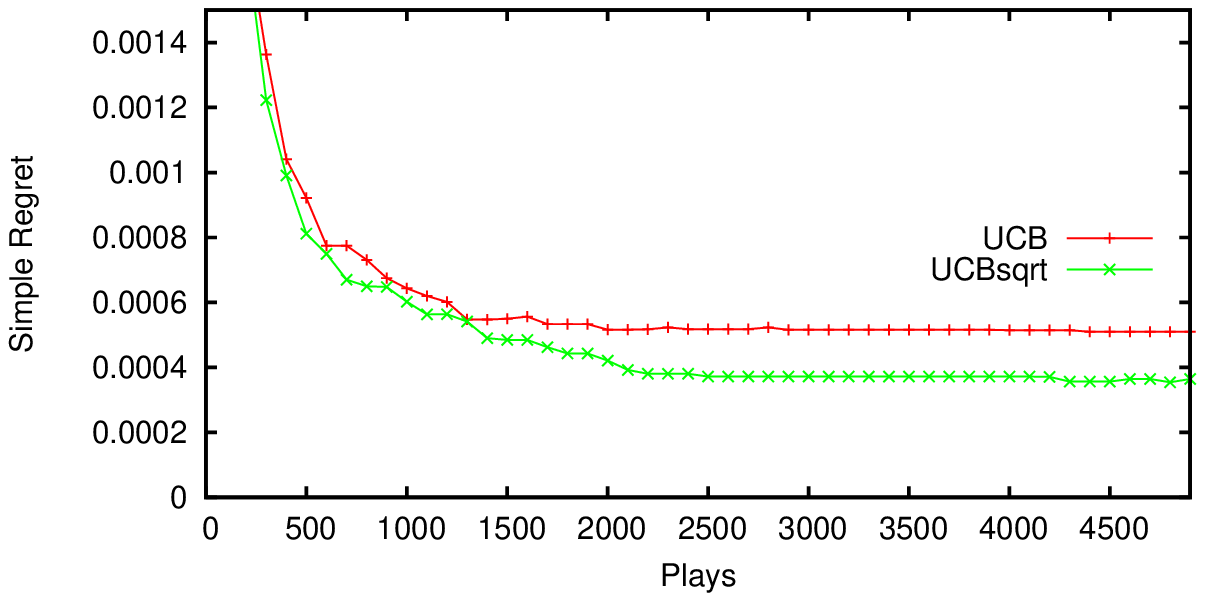}
        \caption{descending reward}
        \label{fig:dsr}
    \end{subfigure}
\end{figure}

The {\it Asymmetric-MCTS} algorithm, which is shown in Algorithm \ref{NodeMCTS}, still retains the four steps in conventional MCTS, namely {\it selection}, {\it expansion}, {\it simulation}, and {\it backpropagation}. The main characteristic of the Asymmetric-MCTS is that it applies the UCB$_{\sqrt{\cdot}}$ algorithm, which is a simple regret bandit algorithm, on max nodes, and the UCB algorithm, which is a cumulative regret bandit algorithm, on min nodes, as shown in Figure \ref{fig:nodewise}.

\section{Experimental Results}

In this section, we will first demonstrate the effect of biased reward on the UCB and UCB$_{\sqrt{\cdot}}$ algorithm. We will then proceed to demonstrate the performance of the Asymmetric-MCTS algorithm on the game of $9\times 9$ Go, $9\times 9$ Nogo, and Othello. The baseline for all experiments is the plain UCT algorithm. For a direct comparison of the effect of the bandit algorithms, all MCTS algorithms used pure random simulations, and no performance enhancement heuristics were applied. Every experimental result is the average of 2300 games, and each algorithm took turns in playing with Black and White.

\subsection{Effect of Biased Reward in the MAB problem}

We will first demonstrate how bias in the reward affects the performance of the UCB and UCB$_{\sqrt{\cdot}}$ algorithm. In order to enhance the effect of the biased reward, we will examine two extreme cases: the rewards are biased to be produced in ascending order, and descending order.

The MAB problem testbed mainly follows the settings specified in Sutton et al. \cite{book}. The results are the average of $2000$ randomly generated $K$-armed bandit problems, with $K= 20$. A total of $5000$ plays were given to each problem. The rewards of each bandit are first generated from a normal (Gaussian) distribution with the mean $w_i$, $i\in K$, and variance of $1$. The mean $w_i$ of the bandits were randomly selected from a normal distribution with mean $0$ and variance 1. To simulate biased rewards, the rewards are then sorted in ascending order and descending order. 

It can be observed from Figure \ref{fig:aop} and Figure \ref{fig:dop} that regardless of the order in which the rewards are biased, the UCB algorithm has a higher percentage of pulling the optimal arm than the UCB$_{\sqrt{\cdot}}$ algorithm, and hence suggesting the UCB$_{\sqrt{\cdot}}$ algorithm tends to distribute its plays more evenly across the candidates. As a result, The UCB algorithm also has lower cumulative regret than the UCB$_{\sqrt{\cdot}}$ algorithm in both cases, as shown in Figure \ref{fig:acr} and Figure \ref{fig:dcr}.  

However, the UCB$_{\sqrt{\cdot}}$ algorithm has a lower simple regret than the UCB algorithm when the rewards are produced in descending order, as shown in Figure \ref{fig:dsr}. As the UCB$_{\sqrt{\cdot}}$ algorithm performs more exploration, it is able to obtain a better estimation of the mean reward of each candidate, and thus can make more informed recommendations, achieving lower simple regret. On the other hand, the extensive explorations performed by the UCB$_{\sqrt{\cdot}}$ algorithm cause its estimations to be too conservative and pessimistic, and hence lower the quality of the recommendations, as shown in Figure \ref{fig:asr}. 

Therefore, it can be observed that the UCB$_{\sqrt{\cdot}}$ algorithm is more conservative in its estimations, and more resistant to situations where it is more likely to make overly optimistic estimations. One the other hand, the UCB algorithm follows closely the change in the reward with high efficiency. 
 
\subsection{Performance of the Asymmetric-MCTS on $9\times 9$ Go}

We will first investigate the performance of the Asymmetric-MCTS on the game of Go played on the $9\times 9$ board, with the komi of 6.5. 

\subsubsection{Performance of SR+CR scheme}

\begin{table}
\caption{Win Rate of SR+CR scheme \cite{simple} against plain UCT algorithm in $9\times 9$ Go. }
\label{tbl:simpleGo}
\centering
{\small
\begin{tabular}{|c|c|}
  \hline
     $c_s$ & SR+CR Scheme \\
  \hline
                   0.1  &  50.00\% $\pm$ 2.04\%\\
                   0.2  &  51.29\% $\pm$ 2.04\%\\
                   0.3  &  51.80\% $\pm$ 2.04\%\\
                   0.4  &  53.91\% $\pm$ 2.04\%\\
                   0.5  &  53.50\% $\pm$ 2.04\%\\
                   0.6  &  51.19\% $\pm$ 2.04\%\\
                   0.7  &  52.23\% $\pm$ 2.04\%\\
                   0.8  &  51.80\% $\pm$ 2.04\%\\
                   0.9  &  {\bf 54.83\% $\pm$ 2.04\%}\\
  \hline
\end{tabular}
}
\end{table}

For comparison, we demonstrate the performance of SR+CR scheme on the game of $9\times 9$ Go. The SR+CR scheme applies the UCB$_{\sqrt{\cdot}}$ bandit algorithm only on the root node, and the UCB bandit algorithm on all other nodes \cite{simple}.  Table \ref{tbl:simpleGo} shows the win rate of various settings for the constant $c_s$ in the UCB$_{\sqrt{\cdot}}$ algorithm in the SR+CR scheme algorithms. The best constant setting for the UCB algorithm is $c_r = 0.4$ in the SR+CR scheme and the plain UCT algorithm is $c = 0.4$. A total of 5000 playouts are given to both algorithms for each move. 

It can be observed that the SR+CR scheme achieves around 54\% with its best constant setting, which is slightly better than the plain UCT algorithm.

\subsubsection{Tuning the C constants}

\begin{table}
\caption{The win rate of the Asymmetric-MCTS with $c_r = 0.5, c_s = 0.4$ against plain UCT algorithm with various constant $c$ settings on $9\times 9$ Go. The optimal setting for plain UCT algorithm is $c = 0.3$, which can only restrict the win rate of Asymmetric-MCTS to 57.70\%. }
\label{tbl:tuneGo}
\centering
{\small
\begin{tabular}{|c|c|}
  \hline
       $c$ & Win Rate    \\
  \hline
      $0.1$ & 58.96\% $\pm$ 2.01\%  \\
      $0.2$ & 59.52\% $\pm$ 2.01\%  \\
   	  $0.3$ & {\bf 57.70\% $\pm$ 2.02\%}  \\
   	  $0.4$ & 58.61\% $\pm$ 2.01\%  \\
   	  $0.5$ & 58.30\% $\pm$ 2.01\%  \\
   	  $0.6$ & 60.74\% $\pm$ 2.00\%  \\
   	  $0.7$ & 62.30\% $\pm$ 1.98\%  \\
   	  $0.8$ & 61.91\% $\pm$ 1.99\%  \\
   	  $0.9$ & 61.61\% $\pm$ 1.99\%  \\
   \hline
\end{tabular}
}
\end{table}

We now proceed to find the best settings for the constant $c_r$ in the UCB algorithm applied on min nodes, and the constant $c_s$ in the the UCB$_{\sqrt{\cdot}}$ algorithm applied on the max nodes, in the Asymmetric-MCTS algorithm. We have found the optimal setting as $c_r = 0.5$ and $c_s = 0.4$, and Table \ref{tbl:tuneGo} shows the win rate of the Asymmetric-MCTS against various constant settings for the plain UCT algorithm. A total of 5000 playouts are given to both algorithms for each move. 

It can be observed that even against the best setting $c=0.3$ of the plain UCT algorithm, the Asymmetric-MCTS still manages to achieve a win rate of around 57.70\%. In comparison to the performance of SR+CR scheme, this result suggests that applying the UCB$_{\sqrt{\cdot}}$ algorithm on the max nodes throughout the game tree, instead of only on the root node, can make a difference.      

\subsubsection{Scalability of Asymmetric-MCTS}

\begin{table}
\caption{Scalability of the Asymmetric-MCTS on $9\times 9$ Go.} 
\label{tbl:scaleGo}
\centering
{\small
\begin{tabular}{|c|c|}
  \hline
     Playouts & Win Rate  \\
  \hline
   1000   & 65.22\% $\pm$ 1.95\% \\
   3000   & 60.00\% $\pm$ 2.00\% \\
   5000   & 57.70\% $\pm$ 2.02\% \\
   7000   & 59.39\% $\pm$ 2.01\% \\
   9000   & 59.57\% $\pm$ 2.01\% \\
   11000  & 62.61\% $\pm$ 1.98\% \\
   \hline
\end{tabular}
}
\end{table}

We now investigate the scalability of the Asymmetric-MCTS as the total number of playouts increases. The result is shown in Table \ref{tbl:scaleGo}.  The settings for Asymmetric-MCTS is  $c_r = 0.5$ and $c_s = 0.4$, and that for the plain UCT algorithm is set to $c=0.3$.

We can observe that the Asymmetric-MCTS achieves a very good win rate of around 65\% over the plain UCT algorithm when 1000 playouts are given, and keeps the win rate to around 60\% as more playouts are given to both algorithms. The results suggest that the Asymmetric-MCTS algorithm has very steady performance on the game of $9\times 9$ Go. 

\subsection{Performance of the Asymmetric-MCTS on $9\times 9$ NoGo}

We now demonstrate the performance of the Asymmetric-MCTS on the game of Nogo. Nogo is a misere variation of the game of Go, in which the first player who has no legal moves other than capturing the stones of the opponent loses. 

\subsubsection{Performance of SR+CR scheme}

\begin{table}
\caption{Win Rate of $UCB_{\sqrt{\cdot}}$ MCTS and SR+CR scheme \cite{simple} against plain UCT algorithm in $9\times 9$ NoGo. The SR+CR scheme has a best win rate of 67.21\% when $c_s = 0.9$ and $c_r = 0.3$.} 
\label{tbl:simpleNoGo}
\centering
{\small
\begin{tabular}{|c|c|}
  \hline
     $c_s$ & SR+CR Scheme \\
  \hline
                   0.1  &   50.23\% $\pm$ 2.04\%  \\
                   0.2  &   51.80\% $\pm$ 2.04\%  \\
                   0.3  &   52.70\% $\pm$ 2.04\%  \\
                   0.4  &   59.60\% $\pm$ 2.01\%  \\
                   0.5  &   64.35\% $\pm$ 1.96\%  \\
                   0.6  &   65.63\% $\pm$ 1.94\%  \\
                   0.7  &   65.28\% $\pm$ 1.95\%  \\
                   0.8  &   66.53\% $\pm$ 1.93\%  \\
                   0.9  &   {\bf 67.21\% $\pm$ 1.92\%}  \\
  \hline
\end{tabular}
}
\end{table}

As in $9\times 9$ Go, we first demonstrate the performance of the SR+CR scheme on the game of $9\times 9$ NoGo for comparison. Table \ref{tbl:simpleNoGo} shows the win rate of various settings for the constant $c_s$ in the SR+CR scheme. The constant setting for the plain UCT algorithm is $c=0.3$. A total of 5000 playouts are given to both algorithms for each move. 

We can observe that SR+CR scheme did extremely well against the plain UCT algorithm, achieving a near 68\% win rate against the plain UCT algorithm.

\subsubsection{Tuning the C constants}

\begin{table}
\caption{The win rate of the Asymmetric-MCTS with $c_r = 0.5, c_s = 0.4$ against plain UCT algorithm with various constant $c$ settings on $9\times 9$ NoGo. The optimal setting for plain UCT algorithm is $c = 0.4$, which can only restrict the win rate of Asymmetric-MCTS to 62.43\%. }
\label{tbl:tuneNoGo}
\centering
{\small
\begin{tabular}{|c|c|}
  \hline
       $c$ & Win Rate    \\
  \hline
      $0.1$ & 64.74\% $\pm$ 1.95\%  \\
      $0.2$ & 66.48\% $\pm$ 1.93\%  \\
   	  $0.3$ & 66.17\% $\pm$ 1.93\%  \\
   	  $0.4$ & {\bf 62.43\% $\pm$ 1.98\%}  \\
   	  $0.5$ & 65.65\% $\pm$ 1.94\%  \\
   	  $0.6$ & 67.00\% $\pm$ 1.92\%  \\
   	  $0.7$ & 67.65\% $\pm$ 1.91\%  \\
   	  $0.8$ & 67.83\% $\pm$ 1.91\%  \\
   	  $0.9$ & 69.83\% $\pm$ 1.88\%  \\
   \hline
\end{tabular}
}
\end{table}

We now proceed to find the best settings for the constants $c_r$ and $c_s$ the UCB$_{\sqrt{\cdot}}$ in the Asymmetric-MCTS algorithm. The optimal setting for the Asymmetric-MCTS algorithm is $c_r = 0.5$ and $c_s = 0.4$. Table \ref{tbl:tuneNoGo} shows the win rate of the Asymmetric-MCTS against various constant settings for the plain UCT algorithm. A total of 5000 playouts are given to both algorithms for each move. 

It can be observed that the Asymmetric-MCTS algorithm achieves at least a win rate of 62.43\% against the plain UCT algorithm. This result suggests that differentiating max nodes and min nodes also produces very good performance, although the SR+CR scheme might be a better choice on the game of $9\times 9$ NoGo. 

\subsubsection{Scalability of Asymmetric-MCTS}

\begin{table}
\caption{Scalability of the Asymmetric-MCTS on $9\times 9$ NoGo.} 
\label{tbl:scaleNoGo}
\centering
{\small
\begin{tabular}{|c|c|}
  \hline
     Playouts & Win Rate  \\
  \hline
   1000   & 57.57\% $\pm$ 2.02\% \\
   3000   & 59.48\% $\pm$ 2.01\% \\
   5000   & 62.43\% $\pm$ 1.98\% \\
   7000   & 65.65\% $\pm$ 1.94\% \\
   9000   & 64.96\% $\pm$ 1.95\% \\
   11000  & 65.96\% $\pm$ 1.94\% \\
   \hline
\end{tabular}
}
\end{table}

We now investigate the scalability of the Asymmetric-MCTS as the total number of playouts increases when applied on $9\times 9$ Nogo. The results are shown in Table \ref{tbl:scaleNoGo}. The settings for Asymmetric-MCTS is  $c_r = 0.5$ and $c_s = 0.4$, and the constant for the plain UCT algorithm is set to $c=0.4$.

It can be observed that the Asymmetric-MCTS algorithm dominates the plain UCT algorithm from a total of 1000 playouts to 11000 playouts, and the win rate gradually increases to near 66\% when 11000 playouts are given to both algorithms. This result suggests that the effect of differentiating max nodes and min nodes will gradually increase with the number of total playouts.

\subsection{Performance of the Asymmetric-MCTS on Othello}

Finally, we proceed to demonstrate the performance of the Asymmetric-MCTS algorithm on the game of Othello.  

\subsubsection{Performance of SR+CR scheme}

\begin{table}
\caption{Win Rate of the SR+CR scheme \cite{simple} against plain UCT algorithm on Othello. The SR+CR scheme has a best win rate of 53.87\%} 
\label{tbl:simpleOthello}
\centering
{\small
\begin{tabular}{|c|c|c|}
  \hline
     $c_s$ & SR+CR Scheme \\
  \hline
                   0.1  &   37.74\%  $\pm$ 1.98\%    \\
                   0.2  &   51.34\%  $\pm$ 2.04\%    \\
                   0.3  &   53.26\%  $\pm$ 2.04\%    \\
                   0.4  &   {\bf 53.87\%  $\pm$ 2.04\%}    \\
                   0.5  &   52.35\%  $\pm$ 2.04\%    \\
                   0.6  &   50.74\%  $\pm$ 2.04\%    \\
                   0.7  &   50.30\%  $\pm$ 2.04\%    \\
                   0.8  &   49.43\%  $\pm$ 2.04\%    \\
                   0.9  &   49.78\%  $\pm$ 2.04\%    \\
  \hline
\end{tabular}
}
\end{table}

We will first investigate the performance of the SR+CR scheme on Othello for comparison. Table \ref{tbl:simpleOthello} shows the win rate of various settings for the constant $c_s$ in the UCB$_{\sqrt{\cdot}}$ algorithm of the SR+CR scheme. The constant setting for the UCB algorithm in the SR+CR scheme is $c_r = 0.6$ and the plain UCT algorithm is $c = 0.6$.  A total of 5000 playouts are given to both algorithms for each move. 

It can be observed that the SR+CR scheme can produce a best win rate of around 53\%, which is slightly better but still around the same level of the plain UCT algorithm.  

\subsubsection{Tuning the C constants}

\begin{table}
\caption{The win rate of the Asymmetric-MCTS with $c_r = 0.7, c_s = 0.4$ against plain UCT algorithm with various constant $c$ settings on Othello. The optimal setting for plain UCT algorithm is $c = 0.6$, which the Asymmetric can only achieve win rate of 50.47\%.}
\label{tbl:tuneOthello}
\centering
{\small
\begin{tabular}{|c|c|}
  \hline
       $c$ & Win Rate    \\
  \hline
      $0.1$ & 88.86\% $\pm$ 1.29\%  \\
      $0.2$ & 81.74\% $\pm$ 1.58\%  \\
   	  $0.3$ & 70.48\% $\pm$ 1.86\%  \\
   	  $0.4$ & 57.61\% $\pm$ 2.02\% \\
   	  $0.5$ & 53.87\% $\pm$ 2.04\%  \\
   	  $0.6$ & {\bf 50.47\% $\pm$ 2.04\%}  \\
   	  $0.7$ & 53.39\% $\pm$ 2.04\%  \\
   	  $0.8$ & 52.13\% $\pm$ 2.04\%  \\
   	  $0.9$ & 53.22\% $\pm$ 2.04\%  \\
   \hline
\end{tabular}
}
\end{table}

We will now proceed to find the best settings for the constants $c_r$ and $c_s$ the UCB$_{\sqrt{\cdot}}$ in the Asymmetric-MCTS algorithm. The optimal setting for the Asymmetric-MCTS algorithm is $c_r = 0.7$ and $c_s = 0.4$. Table \ref{tbl:tuneNoGo} shows the win rate of Asymmetric-MCTS against various constant settings for the plain UCT algorithm. A total of 5000 playouts are given to both algorithms for each move.

It can be observed that the Asymmetric-MCTS algorithm can only achieve a win rate of around 50\% against the plain UCT algorithm. This result suggests that differentiating max nodes and min nodes is not effective on the game of Othello, and is around the same level of performance as the plain UCT algorithm. 

\subsubsection{Scalability of Asymmetric-MCTS}

\begin{table}
\caption{Scalability of the Asymmetric-MCTS on Othello.} 
\label{tbl:scaleOthello}
\centering
{\small
\begin{tabular}{|c|c|}
  \hline
     Playouts & Win Rate  \\
  \hline
   1000   & 52.37\% $\pm$ 2.04\% \\
   3000   & 52.04\% $\pm$ 2.04\% \\
   5000   & 53.43\% $\pm$ 2.04\% \\
   7000   & 50.87\% $\pm$ 2.04\% \\
   9000   & 51.22\% $\pm$ 2.04\% \\
   11000  & 53.43\% $\pm$ 2.04\% \\
   \hline
\end{tabular}
}
\end{table}

We will now investigate the scalability of the Asymmetric-MCTS as the total number of playouts increases when applied on Othello. The results are shown in Table \ref{tbl:scaleOthello}. The settings for Asymmetric-MCTS is  $c_r = 0.7$ and $c_s = 0.4$, and the plain UCT algorithm is set to $c=0.4$.

It can be observed that the performance of the Asymmetric-MCTS algorithm does not change with the increase of the number of playouts. The win rate of Asymmetric-MCTS algorithm holds steady around 50\%, which is around the same performance level as the plain UCT algorithm.

\section{Conclusion}

MCTS has made quite an impact on various fields, and the key to its success lies in the application of bandit algorithms, which solve the MAB problem. In most MCTS variants, the same bandit algorithm and heuristics are applied to every node in the game tree by viewing each node as an independent instance of the MAB problem. The current most dominate variant of MCTS is the UCT algorithm, which applies the UCB bandit algorithm on every node. Although this paradigm has the advantage of allowing MCTS to be applied in a wide spectrum of fields, it leaves a number of properties of the game tree unexploited.

In this work, we have proposed that max nodes and min nodes should be treated differently by applying different bandit algorithms according to its intrinsic nature, rather than using the same bandit algorithm throughout the whole tree. We have observed that different node types have different concerns in their estimation value, and the simple regret bandit algorithms seem to fit the requirements of max nodes, and cumulative regret bandit algorithms seem to fulfill the requirement of min nodes.  

The Asymmetric-MCTS algorithm, which applies the UCB$_{\sqrt{\cdot}}$ algorithm on max nodes, and the UCB algorithm on min nodes is proposed based on this observation. The experimental results show that the Asymmetric-MCTS algorithm has a really good performance and scalability on the games of $9\times 9$ Go. The Asymmetric-MCTS also did well on the game of $9\times 9$ NoGo, but the SR+CR scheme seems to be a better choice. However, the Asymmetric-MCTS performed only on par with the UCT algorithm on the game of Othello. 


As the main difference between the Asymmetric-MCTS algorithm and the UCT algorithm lies in the application of the UCB$_{\sqrt{\cdot}}$ algorithm on max nodes, and hence the effectiveness of the Asymmetric-MCTS algorithm seems to depend on whether the UCB algorithm is more likely to be too optimistic in its estimations on max nodes. Therefore, it can be suggested from the experimental results that the UCB algorithm may make too optimistic estimations on max nodes in the game of $9\times 9$ Go, and on the root node in the game of $9\times 9$ Nogo. On the other hand, situations where the UCB algorithm is likely to be too optimistic rarely occurs in Othello.  

Applying bandit algorithms other than the UCB and the UCB$_{\sqrt{\cdot}}$ algorithm would be a natural direction for further investigation. Apart from bandit algorithms, most performance enhancement methods and heuristics in MCTS, also treats each node in the game tree as equal \cite{sb}\cite{mm}\cite{alphago}. Therefore, it would be interesting to further investigate the possibility of developing enhancement heuristics according to node types as well.






%

\end{document}